%% file: conference_101719.tex
\documentclass[conference,a4paper]{IEEEtran}
\IEEEoverridecommandlockouts

\usepackage{cite}
\usepackage{amsmath,amssymb,amsfonts}
\usepackage{lipsum}
\usepackage{mathtools}
\usepackage{cuted}
\usepackage{algorithm}
\usepackage{algorithmic}
\usepackage{graphicx}
\usepackage{textcomp}
\usepackage{xcolor}
\usepackage{multirow}
\def\BibTeX{{\rm B\kern-.05em{\sc i\kern-.025em b}\kern-.08em
    T\kern-.1667em\lower.7ex\hbox{E}\kern-.125emX}}
\usepackage{etoolbox}
\include{misc/user_colors}
\makeatletter
\patchcmd{\@makecaption}
  {\scshape}
  {}
  {}
  {}
\makeatother

\begin{document}

\title{Reducing The Mismatch Between Marginal and Learned Distributions in Neural Video Compression}
\author{
\IEEEauthorblockN{Muhammet Balcilar}
\IEEEauthorblockA{\textit{InterDigital, Inc.}\\
Rennes, France \\
}
\and
\IEEEauthorblockN{Bharath Bhushan Damodaran}
\IEEEauthorblockA{\textit{InterDigital, Inc.}\\
Rennes, France \\
}
\and
\IEEEauthorblockN{Pierre Hellier}
\IEEEauthorblockA{\textit{InterDigital, Inc.}\\
Rennes, France \\
}
}
\IEEEoverridecommandlockouts
\IEEEpubid{\makebox[\columnwidth]{978-1-6654-7592-1/22/\$31.00~\copyright2022 IEEE\hfill} \hspace{\columnsep}\makebox[\columnwidth]{ }}
\maketitle
\IEEEpubidadjcol
\begin{abstract}
During the last four years, we have witnessed the success of end-to-end trainable models for image compression. Compared to decades of incremental work, these machine learning (ML) techniques learn all the components of the compression technique, which explains their actual superiority. However, end-to-end ML models have not yet reached the performance of traditional video codecs such as VVC. Possible explanations can be put forward: lack of data to account for the temporal redundancy, or inefficiency of latent's density estimation in the neural model. The latter problem can be defined by the discrepancy between the latent's marginal distribution and the learned prior distribution. This mismatch, known as amortization gap of entropy model, enlarges the file size of compressed data. In this paper, we propose to evaluate the amortization gap for three state-of-the-art ML video compression methods. Second, we propose an efficient and generic method to solve the amortization gap and show that it leads to an improvement between $2\%$ to $5\%$ without impacting  reconstruction quality. 

\end{abstract}

\begin{IEEEkeywords}
Neural video compression, Entropy model, Reparameterization.
\end{IEEEkeywords}

\input{introduction}

\input{videocompression}

\input{method}
\input{results}

\bibliographystyle{IEEEtran}
\bibliography{string}


\end{document}

%% file: introduction.tex
\section{Introduction}
\label{sec:intro}
Image and video compression is a fundamental task in image processing, which has become crucial in the time of pandemic and increasing video streaming. Thanks to the community's huge efforts for decades, traditional methods (including linear transformations under heavily optimized handcrafted techniques) have reached current state of the art rate-distortion (RD) performance and dominate current industrial codecs solutions. Alternatively, end-to-end trainable deep models have recently emerged, with promising results. Even though these methods clearly exceed many traditional techniques and surpass human capability for some tasks for a few years back 
very recently they beat the best traditional compressing method (VVC, versatile video coding \cite{vvc}) even in terms of peak signal-to-noise ratio (PSNR) for single image compression \cite{gao2021neural}. However, their performance on video compression are still far from VVC, just on par with one generation back of the traditional method (HEVC, high efficiency video coding \cite{hevc}). In addition to the inefficiency of neural models on capturing temporal redundancy, the mismatch between test latent's normalized histograms and learned distributions in the entropy models may be a contributing factor. 

Lossy image compression via end-to-end trainable models is special kind of Variational Autoencoder (VAE) that learns the transformations among data and latent codes and the probability models of these latent codes jointly \cite{balle2016end,balle2018variational,minen_joint,cheng2020image,gao2021neural}. The problem is a multi-objective optimization problem where the model should be optimized for reconstruction quality and cross entropy of latent code w.r.t learned probabilities known as RD loss function. These neural image codecs were extended by using two VAEs, one for encoding motion information, another for encoding residual information in end-to-end video compression \cite{dvc,ssf,aivc,lhbdc,pourreza2021extending}. As all trainable models suffer from amortization gap \cite{cremer2018inference} (may be optimal for entire dataset, but sub-optimal for given single test instance), neural compression models have a similar issue and this gap reduces the performance by either enlarging the file size or decreasing the reconstruction quality. The first fold solution to this problem is to apply post training for a given single test image/video, where some of them train just the encoder component of VAE \cite{LuCZCOXG20,Campos_2019_CVPR_Workshops} in order to prevent extra signaling cost. The other selection is to fine tune all parts of the model by adding signaling cost to the loss function \cite{van2021overfitting}. Second fold solutions do not apply time-consuming post training, but adjust some parameters of the model. For instance, in \cite{reducinggap}, only the entropy model's amortization gap in end-to-end image compression is targeted and instance specific reparameterization of the latent distribution was proposed. 

This paper is an extension of our previous work \cite{reducinggap} to the video compression with some important differences. First, we introduce a general framework that generalizes end-to-end trainable models on video compression. Second, we analyse the amortization gap of all entropy models for different frames (I, B and P frames), for different information (motion and residual) in three recent neural video compression methods. Third, we identify the origin of the main performance drop and we show how it can be fixed. Last, but not least, we show the efficiency of our probability reparameterization method where the new parameters are kept into file considering the temporal redundancy of these parameters. According to the result, we decrease the file size of video between $2\%$ to $5\%$ in average without any effect on reconstruction quality. To the best of our knowledge, it is the first research on closing the amortization gap of neural video compression without post-training.

%% file: videocompression.tex
\section{End-to-End Video Compression}
\label{sec:vidcomp}

Although the first end-to-end image compression model used only factorized entropy model in \cite{balle2016end}, following papers with hierarchical VAE based hyperprior entropy models as in \cite{balle2018variational,minen_joint,cheng2020image,gao2021neural} became standard in neural image compression and now form the backbone of sota neural video compression. Let $\mathbf{x},\mathbf{\bar{x}},\mathbf{\hat{x}},\mathbf{\hat{x}}_r$ 
 respectively denote the current image, motion warped current image, the reconstruction of current image and the reconstruction of reference image. 
Let us note $\mathbf{v}$ as the predicted motion information, $\mathbf{y}_m,\mathbf{\hat{y}}_m,\mathbf{y}_r,\mathbf{\hat{y}}_r $ 
continuous latent,  quantized (or noise added) latent of motion information and residual information respectively. Similarly, we will note $\mathbf{z}_m,\mathbf{\hat{z}}_m,\mathbf{z}_r,\mathbf{\hat{z}}_r $ 
as the continuous side latent and quantized (or noise added) side latent of motion information and residual information respectively. $\mathbf{Q}(.)$ element-wise function applies quantization in test time or its continuous relaxation in train time as $\mathbf{Q}(x)=x+\epsilon$ that $\epsilon \sim U(-0.5,0.5)$ and $\mathbf{W}(.,.)$ warps given image w.r.t given motion. First VAE whose aim is to encode motion information  takes current frame and reconstructed reference frame (or in B frame encoding, 2 reference frames) as inputs and finds warped image by $\mathbf{y}_m=gm_a(\mathbf{x},\mathbf{\hat{x}}_r;\mathbf{\phi}_m)$, $\mathbf{\hat{y}}_m=\mathbf{Q}(\mathbf{y}_m)$, $\mathbf{v}=gm_s(\mathbf{\hat{y}_m};\mathbf{\theta}_m)$, $\mathbf{z}_m=hm_a(\mathbf{y}_m;\Phi_m)$, $\mathbf{\hat{z}}_m=\mathbf{Q}(\mathbf{z}_m)$,  
and $\mathbf{\bar{x}}=\mathbf{W}(\mathbf{\hat{x}}_r,\mathbf{v})$. Second VAE is for encoding residual information (difference between current frame and its motion warped version) and reconstruction of the image by applying $\mathbf{y}_r=gr_a(\mathbf{x}-\mathbf{\bar{x}};\mathbf{\phi}_r)$, $\mathbf{\hat{y}}_r=\mathbf{Q}(\mathbf{y}_r)$, $\mathbf{z}_r=hr_a(\mathbf{y}_r;\Phi_r)$, $\mathbf{\hat{z}}_r=\mathbf{Q}(\mathbf{z}_r)$ and $\mathbf{\hat{x}}=gr_s(\mathbf{\hat{y}_r};\mathbf{\theta}_r)+\mathbf{\bar{x}}$. As long as $gm_a,gm_s,hm_a,gr_a,gr_s,hr_a$ are trainable deep models, neural video compression loss can be written in followings; 
\begin{multline}
 \label{eq:compvid}
 \scriptstyle
   \mathcal{L}=\mathop{\mathbb{E}}_{\substack{\mathbf{x},\mathbf{\hat{x}}_r\sim p_x \\ \epsilon \sim U}}[-log(p^{(m)}_{h}(\mathbf{\hat{y}}_m|\mathbf{\hat{z}}_m,\mathbf{\Theta}_m))  -log(p^{(m)}_{f}(\mathbf{\hat{z}}_m|\Psi_m)) \\
   \scriptstyle -log(p^{(r)}_{h}(\mathbf{\hat{y}}_r|\mathbf{\hat{z}}_r,\mathbf{\Theta}_r)) -log(p^{(r)}_{f}(\mathbf{\hat{z}}_r|\Psi_r)) + \lambda.d(\mathbf{x},\mathbf{\hat{x}})], 
\end{multline}

\noindent where $p^{(m)}_{f}(.|\Psi_m)$ and
$p^{(r)}_{f}(.|\Psi_r)$ are factorized entropy model for motion and residual information, $p^{(m)}_{h}(.|\mathbf{\hat{z},\Theta_m})$ and $p^{(r)}_{h}(.|\mathbf{\hat{z},\Theta_r})$ are hyperprior entropy models for motion and residual information respectively implemented with neural networks, $d(.,.)$ is any distortion loss such as MSE for PSNR metric, $\lambda$ hyperparameter that plays trade-off role between compression ratio and quality. Here the variables to be written into compressed file are motion's main and side information ($\mathbf{\hat{y}}_m,\mathbf{\hat{z}}_m$) and residual's main and side information ($\mathbf{\hat{y}}_r,\mathbf{\hat{z}}_r$) whose expected file sizes under learned entropy models are the first four parts of the loss in \eqref{eq:compvid}. 

Factorized entropy models learn the probability mass function (pmf) of the symbols for each feature band of $\mathbf{\hat{z}}_m$ and $\mathbf{\hat{z}}_r$ separately. Thus, learned pmf values are defined by parameters of factorized entropy model $\Psi_m,\Psi_r$. If the side latent of motion or residual information is $\mathbf{\hat{z}}_{m|r} \in \mathbf{R}^{k\times k \times f}$ and $\hat{p}_{\Psi_{m|r}}^{(c)}(.)$ is the pmf table of $c$-th feature band for motion or residual information, the factorized entropy model applies;
\begin{equation}
   \label{eq:factorent}
   p^{(m|r)}_{f}(\mathbf{\hat{z}}_{m|r}| \Psi_{m|r})=\prod_{c=1}^{f} \prod_{i,j=1}^{k,k} \hat{p}_{\Psi_{m|r}}^{(c)}({\mathbf{\hat{z}}_{m|r,i,j,c}}).
\end{equation}
Hyperprior entropy model learns the parameters of main information's probability function (usually gaussian or laplacian distribution is used) using already encoded side latent by $\mu,\sigma=hm_s(\mathbf{\hat{z}}_{m};\mathbf{\Theta}_{m})$ or $\mu,\sigma=hr_s(\mathbf{\hat{z}}_{r};\mathbf{\Theta}_{r})$ for motion or residual information and applies as follows;
\begin{equation}
   \label{eq:hyperentapp}
   p^{(m|r)}_{h}(\mathbf{\hat{y}}_{m|r}|\mathbf{\hat{z}}_{m|r},\mathbf{\Theta}_{m|r}) = \prod_{c=1}^{s} \prod_{i \in \mathcal{N}(\sigma_c)} \hat{N}(\mathbf{\Tilde{y}}_{m|r,i};0,\sigma_c),
\end{equation}
\noindent where $\mathbf{\Tilde{y}}_{m|r,i}=Q(\mathbf{y}_{m|r,i}-\mu_i)$, $\mathbf{\hat{y}}_{m|r,i}=\mathbf{\Tilde{y}}_{m|r,i}+\mu_i$, $\sigma_c$ is $c$-th predefined scale, $\mathcal{N}(\sigma_c)$ is a set of latent index whose winning scale is $\sigma_c$, $\hat{N}(x;\mu,\sigma)$ pmf value of $x$ for 1-d gaussian distribution with $\mu,\sigma$ parameters, $s$ is number of predefined scale values of gaussian distribution and $hm_s,hr_s$ are another deep neural networks.

In this framework, end-to-end video compression consists of $5$ trainable components for motion information $gm_a,gm_s,hm_a,hm_s,p^{(m)}_f$ and 5 components for residual information $gr_a,gr_s,hr_a,hr_s,p^{(r)}_f$ parameterized by $\mathbf{\phi}_m,\mathbf{\theta}_m,\Phi_m,\mathbf{\Theta}_m,\Psi_m$ and $\mathbf{\phi}_r,\mathbf{\theta}_r,\Phi_r,\mathbf{\Theta}_r,\Psi_r$ respectively and a non-trainable motion warping function $\mathbf{W}(.,.)$. The selection of all these $11$ components explains the differences between studied end-to-end video compression methods.

%% file: method.tex
\section{Method}
\label{sec:method}
The gap caused by the mismatch between marginal and learned distribution and our proposals are given in this section.
\subsection{Gap of the Entropy Models}
\label{sec:gap}
\begin{figure*}[htb]
\includegraphics[width=0.95\linewidth]{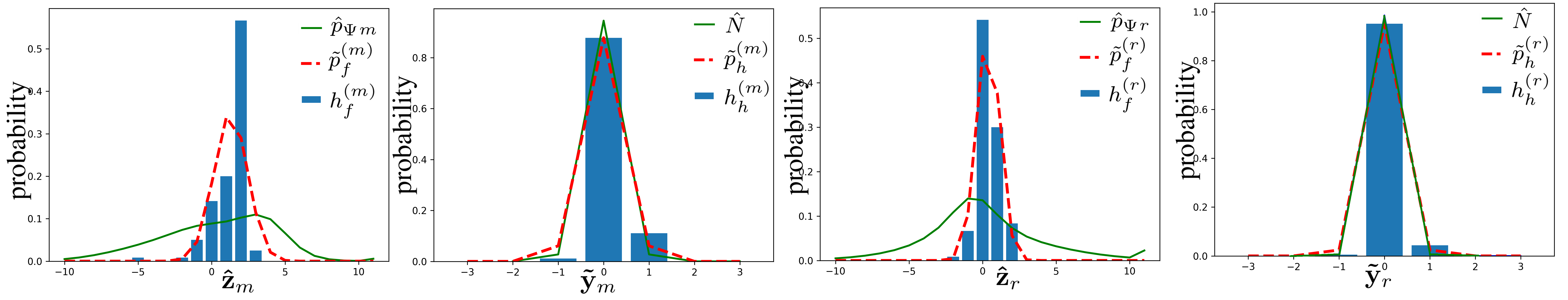}
\caption{Learned pmfs in the model ($\hat{p}_{{\Psi}_m},\hat{N},\hat{p}_{{\Psi}_r},\hat{N}$), pmf after our reparameterization ($\Tilde{p}^{(m)}_f,\Tilde{p}^{(m)}_h,\Tilde{p}^{(r)}_f,\Tilde{p}^{(r)}_h$) and normalized frequencies ($h^{(m)}_f,h^{(m)}_h,h^{(r)}_f,h^{(r)}_h$) for a certain image's selected latents of motion's side, motion's main, residual's side and residual's main information respectively. Our reparameterization fits better on the normalized frequencies, thus it compresses better.}
\label{fig:gap}
\vspace{-0.2cm}
\end{figure*}

When considering intra frame encoding (I frame, as single image encoding) there is no motion nor reference images, thus first two component of the loss in \eqref{eq:compvid} is canceled out and $\mathbf{\bar{x}}=0$. Consequently, only the residual's main and side information have to be encoded for I frames. However, for inter frame encoding (either B or P), both motion's and residual's main and side information are encoded. Using entropy models in \eqref{eq:factorent} and \eqref{eq:hyperentapp}, we can write expected bitlength of $c$-th feature band of side information as follows:
\begin{equation}
   \label{eq:filesize1}
   l^{(m|r)}_{f,c}=-\sum_{i=1}^{k} \sum_{j=1}^{k} log\left(\hat{p}_{\Psi_{m|r}}^{(c)}({\mathbf{\hat{z}}_{m|r,i,j,c}})\right).
\end{equation}
We can write expected bitlength of main information that represented by $c$-th predefined scale as follows;
\begin{equation}
   \label{eq:filesize2}
   l^{(m|r)}_{h,c}=-\sum_{i \in \mathcal{N}(\sigma_c)} log\left(\hat{N}(\mathbf{\Tilde{y}}_{m|r,i};0,\sigma_c)\right).
\end{equation}
Considering all side information's bitlength, it can be calculated by $l^{(m|r)}_f=\sum_c(l^{(m|r)}_{f,c})$ and $l^{(m|r)}_h=\sum_c(l^{(m|r)}_{h,c})$ for all main information's bitlength. This leads us to calculate baseline method's bitlength of inter frame by summing up all 4 informations' bitlength by $l^{(b)}=l^{(m)}_f+l^{(m)}_h+l^{(r)}_f+l^{(r)}_h$. 

The optimality of bit length of each information depends on how learned pmfs ($\hat{p}_{\Psi_m},\hat{p}_{\Psi_r}, \hat{N}(\mathbf{\Tilde{y}}_m;0,\sigma_c), \hat{N}(\mathbf{\Tilde{y}}_r;0,\sigma_c)$ are closed to marginal distribution of latents ($\mathbf{\hat{z}}_m,\mathbf{\hat{z}}_r,\mathbf{\hat{y}}_m,\mathbf{\hat{y}}_r$) and it is defined as amortization gap of entropy models in neural image compression \cite{reducinggap}. This mismatch can be seen by differences between green curves and blue histogram bars in Fig \ref{fig:gap}. Following the same procedure, we can calculate the theoretical limit of expected bit length of each information by simply replacing learned pmf by corresponding latent's normalized histogram as follows;
\begin{equation}
   \label{eq:filesize1lim}
   l^{*(m|r)}_{f,c}=-\sum_{i,j=1}^{k,k} log\left(h_{f,c}^{(m|r)}({\mathbf{\hat{z}}_{m|r,i,j,c}})\right),
\end{equation}
\noindent where $h_{f,c}^{(m|r)}(x)=\sum_{i,j}\delta(x,\mathbf{\hat{z}}_{m|r,i,j,c})/k^2$ is normalized frequency of symbol $x$ on $k \times k$ slice of $\mathbf{\hat{z}}_{m|r}$ and $\delta(.,.)$ is Kronecker delta. Theoretical limit of expected bit length of main information can be written as follows;
\begin{equation}
   \label{eq:filesize2t}
   l^{*(m|r)}_{h,c}=-\sum_{i \in \mathcal{N}(\sigma_c)} log\left(h_{h,c}^{(m|r)}(\mathbf{\Tilde{y}}_{m|r,i})\right),
\end{equation}

\noindent where $h_{h,c}^{(m|r)}(x)=\sum_{i \in \mathcal{N}(\sigma_c)}\delta(x,\mathbf{\Tilde{y}}_{m|r,i})/|\mathcal{N}(\sigma_c)|$ is normalized frequency of symbol $x$ on $\mathbf{\Tilde{y}}_{m|r,i}$ where $i \in \mathcal{N}(\sigma_c)$ and $|\mathcal{N}(\sigma_c)|$ is the number of element in the given set. Following the same procedure, bitlength of all side and main information can be calculated by  $l^{*(m|r)}_f=\sum_c(l^{*(m|r)}_{f,c})$ and $l^{*(m|r)}_h=\sum_c(l^{*(m|r)}_{h,c})$. Thus, theoretical limit of bitlength of inter frame would be $l^{*}=l^{*(m)}_f+l^{*(m)}_h+l^{*(r)}_f+l^{*(r)}_{h}$. The differences between baseline model's each information's bitlength and theoretical limits of the bitlength gives the amortization gap of corresponding type of information.

\begin{algorithm}[tb]
  \caption{Encoding with Factorized Entropy}
  \label{alg:ffe}
  \footnotesize
\begin{algorithmic}
  \STATE {\bfseries Input:} Learned pmfs $\hat{p}_{\Psi}^{(c)}$ for $c=1 \dots s$, latents $\mathbf{\hat{z}} \in \mathbb{R}^{k \times k \times s}$, number of mixture $K$ and previous inter frame's params $\hat{\beta}_{p}^{*(c)}$ for $c=1 \dots s$. 
  \STATE {\bfseries Output:} latent bitstream $lb$ and parameter bitstream $pb$. 
  \STATE Reorder $\hat{p}_{\Psi}^{(c)}$ for $c=1 \dots s$ w.r.t its entropy
  \FOR{$c=1$ {\bfseries to} $S$} \STATE $\beta^{*(c)}=argmax_{\beta} \sum_{i,j} log\left(\Tilde{p}(\mathbf{\hat{z}}_{i,j,c});\beta^{(c)})\right)$ // find parameters
  \STATE $\hat{\beta}^{*(c)}=Q(\beta^{*(c)})$ // quantize the best parameters
  \STATE $\mathcal{G}= \sum_{i,j}-log_2(\hat{p}_{\Psi}^{(c)}(\mathbf{\hat{z}}_{i,j,c}))+log_2(\Tilde{p}(\mathbf{\hat{z}}_{i,j,c};\hat{\beta}^{*(c)}))$ // gain
    \IF{$\hat{\beta}_p^{*(c)}=None$ and $\mathcal{G}<=30K-10$ }
    \STATE $pb.write(True)$ // write temporal bit
    \STATE $\hat{\beta}^{*(c)}=None$ // set best params None
    \STATE $lb.write(\mathbf{\hat{z}}_{:,c},\hat{p}_{\Psi}^{(c)})$ // write latent w.r.t learned pmf
  \ELSIF{$\hat{\beta}^{*(c)}=\hat{\beta}_p^{*(c)}$ and $\mathcal{G}>0$ }
  \STATE $pb.write(True)$ // write temporal bit
  \STATE $lb.write(\mathbf{\hat{z}}_{:,c},\Tilde{p}(.;\hat{\beta}_p^{*(c)}))$ // write latent w.r.t previous pmf
  \ELSIF{$\mathcal{G}>30K-10$}
   \STATE $pb.write(False)$ // write temporal bit
   \STATE $pb.write(True)$   // write replacement bit
   \STATE $pb.write(\hat{\beta}^{*(c)})$ // write parameters explicitly
    \STATE $lb.write(\mathbf{\hat{z}}_{:,c},\Tilde{p}(.;\hat{\beta}^{*(c)}))$ // write latent w.r.t new pmf
  \ELSE
     \STATE $pb.write(False)$ // write temporal bit
   \STATE $pb.write(False)$   // write replacement bit
   \STATE $\hat{\beta}^{*(c)}=None$ // set best params None
    \STATE $lb.write(\mathbf{\hat{z}}_{:,c},\hat{p}_{\Psi}^{(c)})$ // write latent w.r.t learned pmf
  \ENDIF
  \ENDFOR
  \FOR{$c=S+1$ {\bfseries to} $s$} 
  \STATE $lb.write(\mathbf{\hat{z}}_{:,c},\hat{p}_{\Psi}^{(c)})$ // write latent w.r.t learned pmf
  \ENDFOR
\end{algorithmic}
\end{algorithm}

\subsection{Temporal Re-parameterization}
\label{sec:reparam}
In order to reduce the mismatch, learned pmfs in the model were replaced by some parametric distribution $\Tilde{p}_{f|h}^{(m|r)}(.;\beta)$ 
whose parameters $\beta$, to be signalled to the receiver, are obtained by fitting the actual histogram of the latents as in \cite{reducinggap}. We use the same reparameterizations that are truncated Gaussian mixture for factorized entropy (motion/residual's side information) and truncated zero-mean Gaussian distribution for hyperprior entropy (motion/residual's main information) where the parameters are discretized into 10 bits. Sample probabilities with $K=1$ Gaussian mixture and zero-mean Gaussian can be found in Fig. \ref{fig:gap} in the red curves for all four types of information. When $l^{o(m|r)}_{f|h}$ is the bitlength of motion/residual's side/main information under $\Tilde{p}_{f|h}^{(m|r)}(.;\beta)$ probabilities, plus the signaling cost of $\beta$, the bitlength of interframe in our proposal would be $l^{(o)}=l^{o(m)}_f+l^{o(m)}_h+l^{o(r)}_f+l^{o(r)}_{h}$.  
\begin{table*}[htbp]
\caption{Ratio of amount of certain information in the bitstream, its amortization gap and our savings for different frame types and 16 length sequence of video. All numbers are percentage and obtained by average results   of 7 videos in UVG test video set. All methods are tested for the provided lowest bit rate.}
\vspace{-0.2cm}
\begin{center}
\begin{tabular}{|cc|ccc|ccc|ccc|ccc|cc|}
\hline
\multirow{3}{*}{Method} &\multirow{3}{*}{Frame} &\multicolumn{6}{|c|}{Motion Information} & \multicolumn{6}{|c|}{Residual Information} & \multicolumn{2}{|c|}{All} \\
 & &\multicolumn{3}{|c|}{Factorized Entropy (Side)} & \multicolumn{3}{|c|}{Hyperprior Entropy (Main)} & \multicolumn{3}{|c|}{Factorized Entropy (Side)} & \multicolumn{3}{|c|}{Hyperprior Entropy (Main)} & \multicolumn{2}{|c|}{}
 \\
 & & Ratio& Gap & Saving & Ratio& Gap & Saving & Ratio& Gap & Saving & Ratio& Gap & Saving & Gap & Saving\\
\hline
\multirow{3}{*}{SSF \cite{ssf}} &I & - & -  & -  & - & -  & - & 16.1 & 10.6 & 5.5 & 83.9& 1.7 & 0.9 & 3.1 & 1.7\\
 &P & 5.1& 22.1 & 11.7 & 2.6& 6.3 & 0.9 & 15.4& 11.1 & 5.4 & 76.9& 1.5 & 0.9 & 4.2 & 2.2\\
  &Video & 4.4 &  22.1  & 11.7   & 2.3 & 6.3     & 0.9  & 15.5 & 11.0  & 5.4  & 77.8& 1.6  & 0.9  & 4.1  & 2.0 \\
\hline
\multirow{3}{*}{LHBDC \cite{lhbdc}} &I & - & -  & -  & - &  - &  - & 10.2& 14.7 & 11.1 & 89.8& 1.3 & 0.8 & 2.7 & 1.9\\
 &B & 0.4& 18.0 & 1.6 & 3.5& 12.0 & 1.9 & 13.2& 43.9 & 37.2 & 82.9& 2.6 & 0.8 & 8.3 & 5.6\\
  &Video & 0.2& 18.0 & 1.6 & 2.1& 12.0 & 1.9 & 12.2& 33.6 & 27.9 & 85.5& 2.0 & 0.8 & 6.1 & 4.1\\
\hline
\multirow{4}{*}{AIVC \cite{aivc}} &I & - & -  & -  &- &  - &  - & 6.2& 38.5 & 34.1 & 93.8& 1.6 & 0.6 & 3.8 & 2.7\\
 &P & 3.5& 32.7 & 20.0 & 0.4& 17.5 & 0.0 & 8.2& 25.1 & 18.3 & 87.9& 2.3 & 1.0 & 5.3 & 3.0\\
 &B & 3.9& 38.5 & 19.3 & 0.1& 26.0 & 0.0 & 14.9& 40.9 & 27.7 & 81.1& 4.3 & 2.2 & 11.1 & 6.6\\
  &Video & 2.4& 37.5 & 19.4 & 0.1& 22.7 & 0.0 & 10.8& 39.0 & 28.2 & 86.7& 2.9 & 1.4 & 7.7 & 4.7\\
\hline
\end{tabular}
\label{tab1}
\end{center}
\vskip -0.08in
\end{table*}
\begin{figure*}[htb]
\centering
\includegraphics[width=0.75\linewidth]{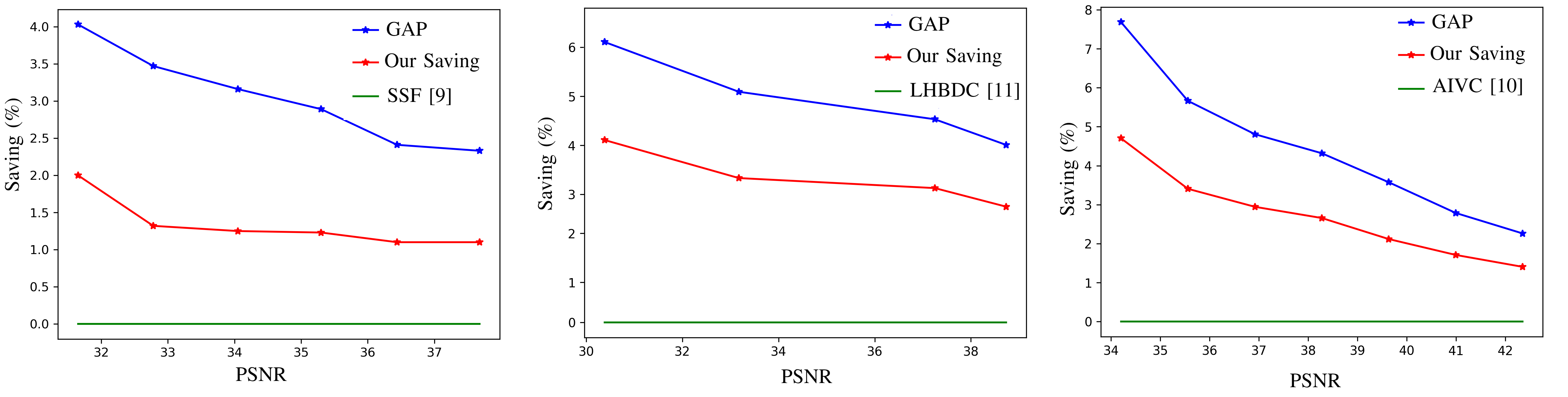}
\caption{Amortization gaps and our savings relative to total file size of 3 recent neural video compression methods for 16 frame length video sequence compression on 7 videos in UVG dataset under different reconstruction quality.}
\label{fig:gapres}
\vskip -0.08in
\end{figure*}
To take temporal redundancies of parameter signaling cost into account, we use $S$-bit temporal mask to explain if the previously encoded interframe's corresponding parameter is the same or not for the top $S$ number of pmf tables. If this bit is 1, there is no need to encode new parameters but uses previously encoded interframe's parameters. In this way, we decrease the temporal redundancy of these parameters. When considering the signaling cost of each pmf table, we have seen that just a few of the pmf table is worth to replace with reparameterization one but vast majority of them is not. To determine which is worth or not, we reorder the pmf table w.r.t its own entropy (in hyperprior model, it is not necessary since the pmf tables are ordered by predefined scale values and lower scales pmf table highly likely carries more information). Later we 
test top $S$ number of pmf table one by one if the expected bit gain is larger than reparameterization cost or not. The new parameters are written to the file and pmf is replaced if the gain is larger than that signaling cost. When $K$ number of mixture is used, the signaling cost is $10(3K-1)$ bits in factorized entropy. Since there is just one parameter in zero-mean gaussian, the signaling cost is $10$ bits in hyperprior model when temporal mask is 0. If the parameters are the same with previously encoded interframe, signaling cost is 0 bit for all entropy models. To tell the receiver which pmf is replaced, we need to use 1-bit replacement mask in addition to replace pmf's parameters and temporal mask. As long as $\Tilde{p}(.;\hat{\beta}^{(c)})$ is the new $c$-th pmf table parameterized by $\hat{\beta}^{(c)}$ and previous encoded interframe's parameter is $\hat{\beta}_p^{*(c)}$ (all params of the first interframe, and the pmf table which are not replaced are $None$), detailed flow of factorized entropy encoding is given in Algorithm~\ref{alg:ffe}. 
Extending it on hyperprior entropy model is straightforward by changing only the variable names/indexing. 

%% file: results.tex
\section{Results}
\label{sec:results}
In order to test our proposal, we use 7 video sequences in UVG dataset \cite{mercat2020uvg} at 1080p resolution. We use 10 set of consecutive 16 frames (total 160 frames) of each sequence and compress them by compressai \cite{compressai} implementation of SSF \cite{ssf}, and author's official implementations of LHBDC \cite{lhbdc} and AIVC \cite{aivc}. Since all methods were proposed on different purpose, we do not make comparison of these methods but measure saving of our generic solution up to these baselines model. SSF encodes first frame as I frame and rest of the 15 frame as P frames. LHBDC needs 2 reference frames. We supposed first frame and 17th frame is I frame and all 15 frames in between frames are B frames encoded by hierarchical bi-directional settings. In calculations, we did not count 17th frame, because it is also next GOP's first reference frame and should be accounted in next GOP's file size. AIVC encodes first frame as I frame, 16th frames as P frames and rest of the in between frames as B frames. In order to show detailed gap sources, 
we give the result in Table \ref{tab1} for the lowest bpp objective. In all cases, the ratio is percentage of each information's bitlength ($l_f^{(m)},l_h^{(m)},l_f^{(r)},l_h^{(r)}$) w.r.t the total bitlength $l^{(b)}$ in the baseline model. The gap of each information is  $1-l_{f|h}^{*(m|r)}/l_{f|h}^{(m|r)}$ while the total gap is $1-l^{*}/l^{(b)}$. It can be seen that for all methods, the gap is not negligible since a potential performance gain of several percent could be achieved. We calculate our savings by $1-l_{f|h}^{o(m|r)}/l_{f|h}^{(m|r)}$ for each information type, where the total saving is  $1-l^{(o)}/l^{(b)}$. We measure those percentages for each frame type as well as 16-length sequences of videos reported on the Table \ref{tab1}. 

In all models and frame types, the residual's main information represents the vast majority of the bitstream ($76.9\%-93.8\%$) whose gap is the smallest ($1.3\%-4.3\%$). Our proposal can fill almost half of this gap. Residual's side information is the second largest information type in the interframe's bitstream where the biggest gap exists ($10.6\%-43.9\%$). Truncated gaussian mixture fills more than half of these gaps. Even though the gap of motion information is larger, the amount of the data is small and its effect on general performance is limited. Since our reparameterization approach has fixed cost, if the amount of data is very low (less than $1\%$ in AIVC and around $2\%$ in SSF's main motion information and less than $1\%$ in LHBDC's side motion information), proposed method cannot save any bits or saving is very limited even though the gap is $10\%-20\%$. Our performance on side information (factorized entropy) is much more larger than performance on main information (hyperprior entropy) for all methods. In average of 16 frame length video for the lowest bpp objective, the gap is between $4.1\%-7.7\%$ and our saving is between $2.0\%-4.7\%$. To measure the gap and our savings in different quality/rate, we perform several tests with the pretrained models provided by authors. Results are given in Fig~\ref{fig:gapres} and show that gap and our savings decreases with the increment of quality/rate. Even at the highest quality/rate objective, the gain is better than $1\%$ for all methods. 

\section{Conclusion}
\label{sec:conc}
In this work, we propose a low computational demanding solution for the inefficiency of entropy coding in neural video compression. 
Since the method is generic, it can be applied up to any neural video compression and decrease the bitlength between $2\%-5\%$ according to our tests. The amortization gap is inevitable in all ML based models. However, the connection of the amount of this gap to the certain type of neural codec should be explored in order to develop better codec which is inside our agenda as a future work.